\documentclass[letterpaper]{article} 
\usepackage{aaai25}  
\usepackage{times}  
\usepackage{helvet}  
\usepackage{courier}  
\usepackage[hyphens]{url}  
\usepackage{graphicx} 
\usepackage{natbib}  
\usepackage{caption} 
\usepackage{algorithm}
\usepackage{algorithmic}
\usepackage{booktabs}
\usepackage{newfloat}
\usepackage{listings}

\usepackage{amsmath,amsfonts,bm}

\def\eqref#1{equation~\ref{#1}}

\def\1{\bm{1}}

\def\mM{{\bm{M}}}

\def\mW{{\bm{W}}}
\def\mX{{\bm{X}}}

\DeclareMathAlphabet{\mathsfit}{\encodingdefault}{\sfdefault}{m}{sl}
\SetMathAlphabet{\mathsfit}{bold}{\encodingdefault}{\sfdefault}{bx}{n}

\def\gD{{\mathcal{D}}}

\def\gL{{\mathcal{L}}}

\def\gO{{\mathcal{O}}}

\DeclareMathOperator*{\argmin}{arg\,min}

\usepackage{color}
\usepackage{amsthm}
\usepackage{pifont}
\usepackage{multirow}
\usepackage{subcaption}

\DeclareCaptionStyle{ruled}{labelfont=normalfont,labelsep=colon,strut=off} 
\floatstyle{ruled}
\newfloat{listing}{tb}{lst}{}
\floatname{listing}{Listing}
\title{Federated Foundation Models on Heterogeneous Time Series}
\author{
    Shengchao Chen\textsuperscript{\ding{169}}, Guodong Long\textsuperscript{\ding{169}}, Jing Jiang\textsuperscript{\ding{169}}, Chengqi Zhang\textsuperscript{\ding{171}}\\
}
\affiliations{
    \textsuperscript{\ding{169}}Australian Artificial Intelligence Institute, University of Technology Sydney\\
    \textsuperscript{\ding{171}}Department of Data Science and Artificial Intelligence, The Hong Kong Polytechnic University \\
    \texttt{shengchao.chen.uts@gmail.com}, \texttt{\{guodong.long, jing.jiang\}@uts.edu.au} \\
    \texttt{chengqi.zhang@polyu.edu.hk}\\
}

\usepackage{bibentry}

\begin{document}

\maketitle

\begin{abstract}
Training a general-purpose time series foundation models with robust generalization capabilities across diverse applications from scratch is still an open challenge. Efforts are primarily focused on fusing cross-domain time series datasets to extract shared subsequences as tokens for training models on Transformer architecture. However, due to significant statistical heterogeneity across domains, this cross-domain fusing approach doesn't work effectively as the same as fusing texts and images. To tackle this challenge, this paper proposes a novel federated learning approach to address the heterogeneity in time series foundation models training, namely FFTS. Specifically, each data-holding organization is treated as an independent client in a collaborative learning framework with federated settings, and then many client-specific local models will be trained to preserve the unique characteristics per dataset. Moreover, a new regularization mechanism will be applied to both client-side and server-side, thus to align the shared knowledge across heterogeneous datasets from different domains. Extensive experiments on benchmark datasets demonstrate the effectiveness of the proposed federated learning approach. The newly learned time series foundation models achieve superior generalization capabilities on cross-domain time series analysis tasks, including forecasting, imputation, and anomaly detection. Code available at: \textcolor{blue}{\texttt{https://github.com/shengchaochen82/FFTS}}
\end{abstract}

\section{Introduction}
Training time series foundation models (TSFMs) requires access to numerous publicly available datasets and a large number of datasets provided by various organizations. Federated learning for Foundation Models~\cite{zhuang2023foundation} is a new pathway to achieve this utilizing both public and private datasets. Existing efforts focus on training TSFMs in a centralized manner, which is unsuitable for accessing datasets in the private domain~\cite{liu2024timer,goswami2024moment,das2023decoder,woo2024unifiedtraininguniversaltime}. For example, cross-silo federated learning~\cite{mcmahan2017communication} explores cross-institutional collaborations among hospitals or financial institutions. Moreover, many end-users with wearable devices and self-driving cars are highly concerned about the use of data collected by these devices.

In comparison to text and images, a unique challenge for time series foundation models is the increasing heterogeneity of tokens. The token, a subsequence of a time series, can represent distinct physical meanings in different application scenarios while the words or image objects usually represent similar meanings in different domains. Lack of cross-domain invariant and increasing heterogeneity of time series significantly impact the performance of foundation models.

\begin{figure}[tbh]
    \centering
    \includegraphics[width=.45\textwidth]{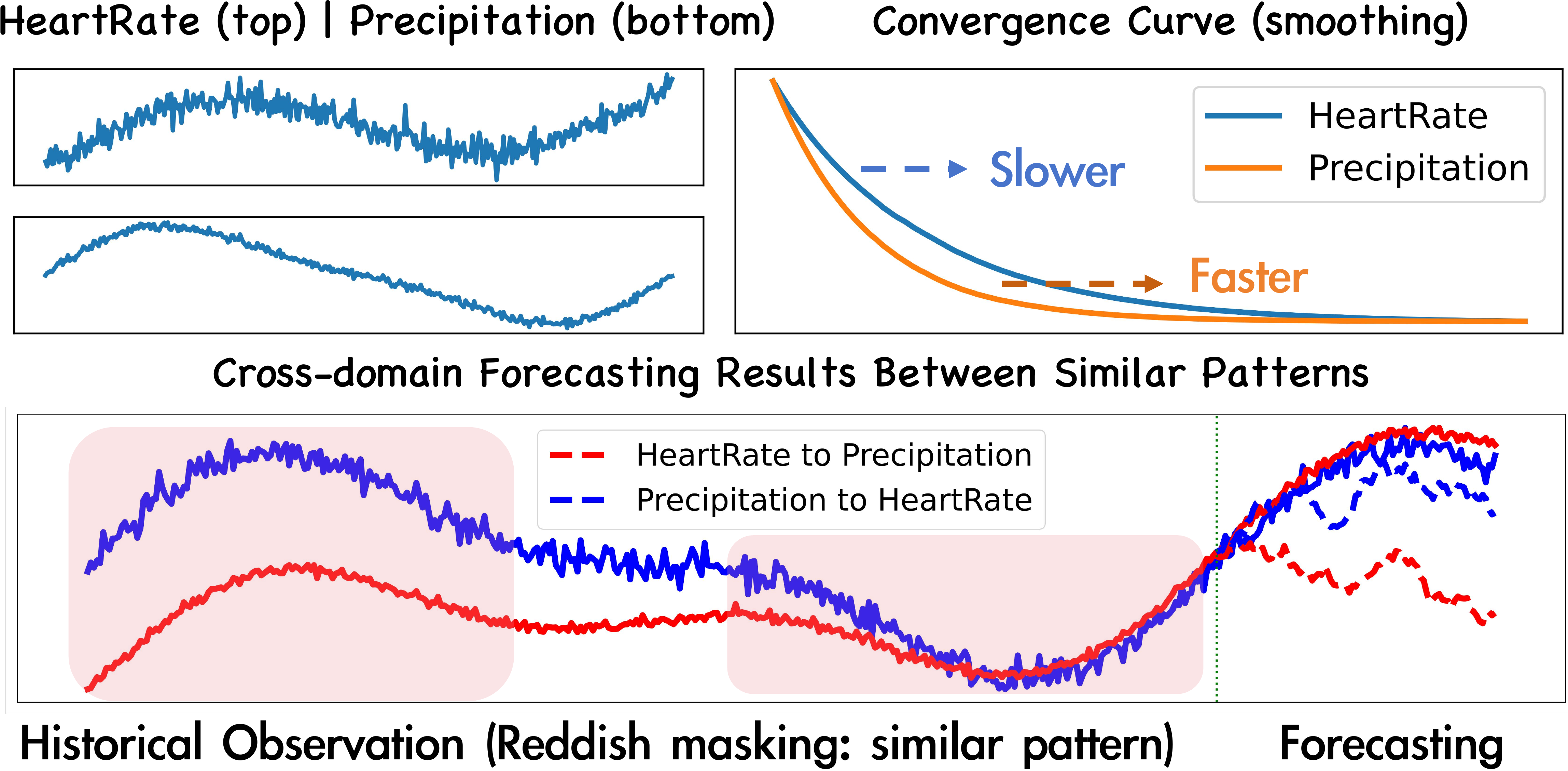}
    \caption{\small Examples of statistical heterogeneity across time series datasets. HeartRate: healthcare data, Precipitation: weather data.}
    \label{fig:example_hete}
\end{figure}

Heterogeneity arises as cross-domain time series often exhibit significant variations in temporal patterns, including trends and timescales. This leads to two main issues: \textbf{(1)} inconsistent convergence rates across different domains (refer to \textbf{Fig.~\ref{fig:example_hete}}, top): models trained on complex healthcare data converge more slowly compared to those trained on simpler weather data, despite similar underlying trends; and \textbf{(2)} inability of models to leverage heterogeneous data with analogous patterns effectively (refer to \textbf{Fig.~\ref{fig:example_hete}}, bottom): models trained on weather data underperform when applied to healthcare data and vice versa, even though both datasets exhibit similar patterns. Additionally, distinct domain-specific contextual meanings within different timescales often imply unique interpretations, such as fluctuations in weather or health status changes, imply unique interpretations. Despite uniform observation periods, these timescale differences can skew meanings and complicate data integration for training, often causing models to misinterpret underlying patterns when trained on fused cross-domain time series.

\begin{figure*}[tbh]
    \centering
    \includegraphics[width=.975\textwidth]{ 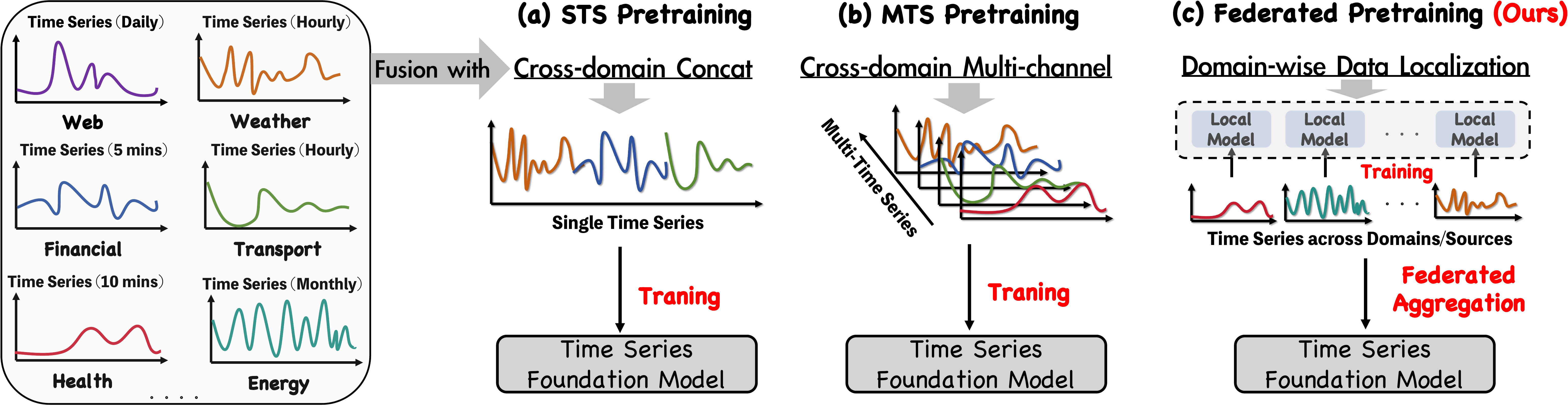}
    \caption{\small \textit{Overview} of Time Series Foundation Models (TSFMs) that training from scratch. (a) \textbf{STS Pretraing:} Training from scratch using a single time series (STS) obtained by fusing data from different domains~\cite{goswami2024moment,liu2024timer,das2023decoder}. (b) \textbf{MTS pretraing:} Training from scratch using multiple time series (MTS) from different domains~\cite{woo2024unifiedtraininguniversaltime}. (c) \textbf{Federated Pretraining (Ours, this paper):} Instead of merging time series from various domains, separate local models are trained for each source. These models are then aggregated into a global model on the server to form a TSFM.}
    \label{fig:approach_overview}
\end{figure*}

This paper presents FFTS, a novel federated learning approach to address the heterogeneity in time series foundation models training. As shown in \textbf{Fig.~\ref{fig:approach_overview}(c)}, FFTS treats each data-holding organization as an independent client within a collaborative framework. Each client trains a local model to preserve unique dataset characteristics, while a server aggregates these to form a TSFM. FFTS provides the insights for training TSFM on heterogeneous time series in two ways: model architecture and optimization. We adopt a standard encoder-only Transformer architecture and introduce an adaptive trend awareness module to identify similar patterns among heterogeneous sequences. For optimization, we use a uniform masking strategy to prevent local models from memorizing domain-specific knowledge and introduce a heterogeneous knowledge alignment strategy to reduce bias in the global model. A unified adaptation architecture supports diverse downstream tasks. FFTS shows robust generalization and outperforms existing task-specific models. Key contributions include:
\begin{itemize}
    \item This is the first exploration of the potential for using FL to train time series foundation models for generalization across various downstream application and tasks.

    \item We present an empirical qualitative analysis demonstrating how statistical heterogeneity across cross-domain time series can negatively affect centralised training.
    
    \item We introduce FFTS, a FL approach to tackle statistical heterogeneity in training time series foundation models. FFTS provides insights from model architecture and optimization to enhance training.

    
    
    \item Extensive experiments on real-world time series datasets demonstrate our FFTS exhibits robust zero/few-shot generalization in long-term forecasting and outperform state-of-the-art task-specific models in forecasting, imputation, and anomaly detection.
\end{itemize}

\section{Related Work}
\paragraph{Time Series Foundation Models} Pre-trained models have evolved into large foundation models, effectively handling diverse data across domains and tasks with notable advancements in few/zero-shot generalization~\cite{chen2023foundation}. However, Time Series Foundation Models (TSFMs) are still underdeveloped due to data heterogeneity and scarcity. Research on TSFMs is emerging and broadly falls into two categories: (1) Adapting existing LLMs to TSFMs, which involves enhancing their forecasting accuracy through fine-tuning~\cite{zhou2023one,chang2023llm4ts} or using multimodal prompts~\cite{jin2023time,liu2024unitime,cao2023tempo}. These approaches depend significantly on the quality of the LLM backbone and effective cross-modal alignment. (2) Training TSFMs from scratch using extensive time series datasets, either real-world~\cite{goswami2024moment,liu2024timer,garza2023timegpt} or synthetic~\cite{dooley2024forecastpfn}. This method, although robust for generalization across various domains, is resource-intensive and faces challenges due to the heterogeneous nature of time series data and privacy concerns related to centralized data training. Our FFTS can train a TSFM from scratch \textit{\textbf{without}} (1) fusing large-scale datasets on the server, and (2) centralized training raises privacy concerns.

\paragraph{Federated Learning in Time Series} Federated Learning (FL) is a distributed learning paradigm that enables collaborative model training while preserving data privacy~\cite{mcmahan2017communication,chen2024free}. It has been effectively applied in real-world time series analyses across various domains including climate change~\cite{chen2023prompt,chen2023federated}, energy~\cite{sun2024hifi} and finance~\cite{nevrataki2023survey}. Motivated by the success of foundation models and heightened privacy concerns~\cite{zhuang2023foundation}, FL is increasingly used to fine-tune pre-trained LLMs, allowing for personalized training that addresses data heterogeneity among clients in distributed time series analysis~\cite{chen2024personalized,liu2024time}. However, the challenge of training a time series foundation model from scratch in a heterogeneous data environment using FL remains a open challenge.


\section{Preliminary}
\paragraph{Problem Definition} 
This paper explores training a TSFM from scratch on heterogeneous time series using FL. \textbf{Fig.~\ref{fig:approach_overview}} illustrates the primary strategies for training TSFMs. These strategies are categorized into three approaches based on data fusion methods: (1) Single Time Series (STS) pre-training involves concatenating multiple datasets along the time dimension to train a model for a single time series, as shwon in \textbf{Fig.~\ref{fig:approach_overview}(a)}. (2) Multiple Time Series (MTS) pre-training combines multiple datasets along the channel dimension to train a model for a multi-channel time series, as shown in \textbf{Fig.~\ref{fig:approach_overview}(b)}. (3) Our proposed Federated Pre-training approach treats each time series as an independent client, training a separate model for each client and then aggregating them globally to generate a TSFM, as shown in \textbf{Fig.~\ref{fig:approach_overview}(c)}. We define an observation of a multi-channel time series from a given domain $d \in \gD$ is as $\mX_d \in \mathbb{R}^{L \times C}$, where $C$ and $L$ represent the number of channels and the observation length, respectively. Each domain (data-holder), viewed as an independent client, updates its own model, $F(\cdot)$ parameterized by $\theta$, to minimize the distance between output and ground truth. The global optimization objective is to create a uniform model by integrating each client's knowledge as:
\begin{equation}
    F(\theta) \text{:}= \argmin \sum_{i=1}^{N} \frac{n_k}{n} F_i(\theta_i; \lbrace D_i \rbrace),
\end{equation}
where $n_i$ and $n$ is the number of samples held by the $k$-th client and all clients, respectively, and $F(\theta; \lbrace D \rbrace)$ denotes the local objective function, such as mean square error (MSE). This paper proposes FFTS, offering insights into training federated foundation models on heterogeneous time series from both model architecture and optimization perspectives. We will elaborate on these aspects.

\section{FFTS: Model Architecture}
\begin{figure}[tbh]
    \centering
    \includegraphics[width=.485\textwidth]{ 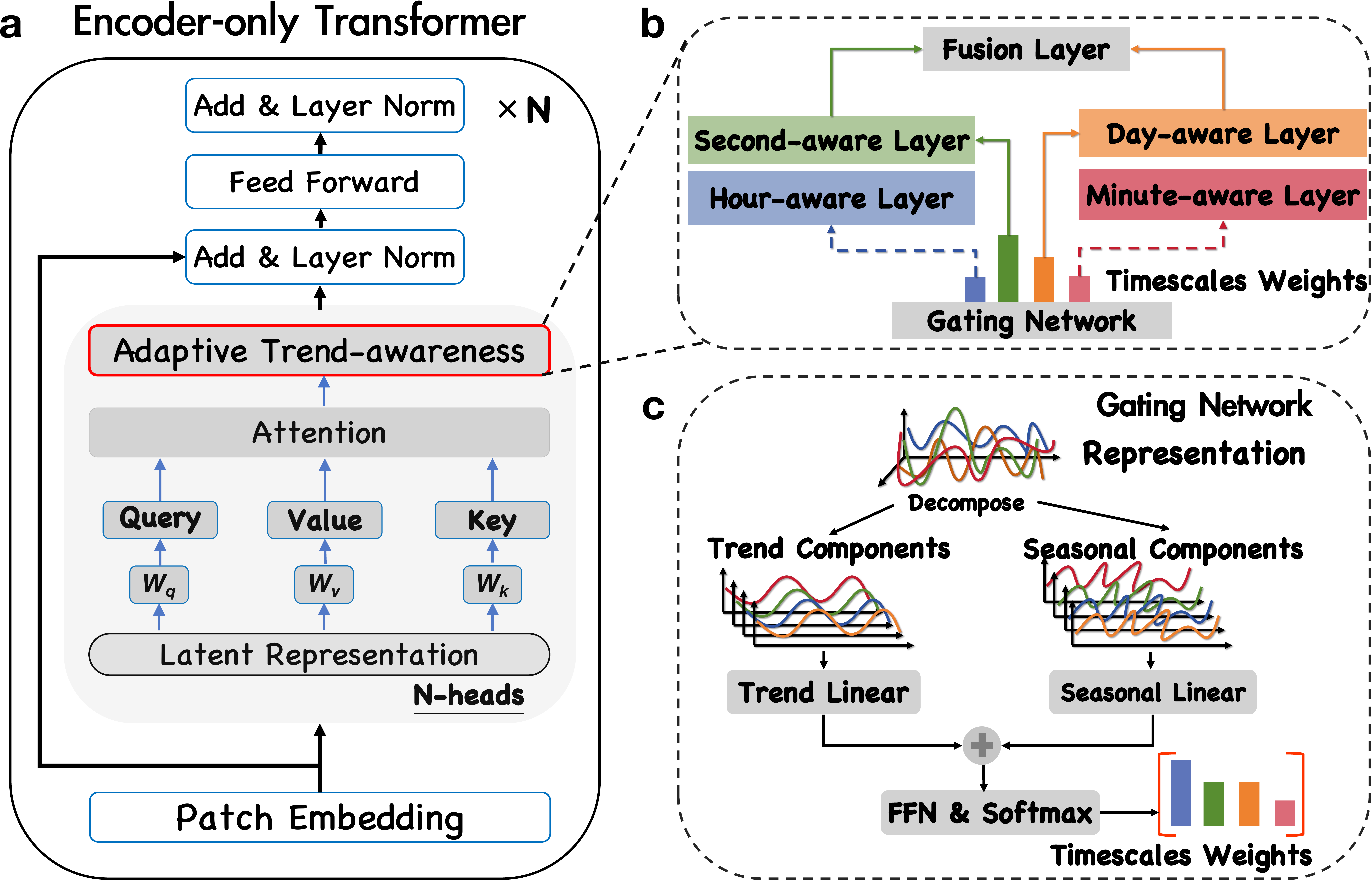}
    \caption{\small \textit{Architecture of model within FFTS}. \textbf{a} Structure of local model in each client. \textbf{b} Architecture of the proposed Adaptive Trend-awareness Module (ATM), which consists of four independent experts for extracting trends at different timescales based on the representation from Attention. Structurally inspired by the Mixture of Experts (MoE)~\cite{fedus2022switch}. \textbf{c} Architecture of the Gating Network.}
    \label{fig:local_architecture}
\end{figure}

As shown in \textbf{Fig.~\ref{fig:approach_overview}(c)}, our FFTS treats each data-holder as an independent client, training their own local model and uploading the local model's parameters to the server to generate a global model. The architecture of the local model on each client is depicted in \textbf{Fig.~\ref{fig:local_architecture}}, which is based on a standard encoder-only Transformer (\textbf{Fig.~\ref{fig:local_architecture}a}) to ensure FFTS's flexibility and versatility. Specifically, we adopt patch embedding to enhance the representation of time series. Additionally, we introduce adaptive trend-awareness module to discover complex temporal patterns in time series and mitigate the influence of heterogeneity across cross-domain data for the global model. Further details are described below.

\paragraph{Patch Embedding} To address the lack of semantic information in individual time points of time series data, we utilize a patching strategy that groups neighboring time points into discrete tokens~\cite{nie2022time}, achieving two main objectives: (1) capturing local semantic patterns, and (2) enhancing computational efficiency for long sequences. Initially, each time series $\mathbf{X}$ is normalized to zero mean and unit variance via Reverse Instance Normalization~\cite{kim2021reversible} to counter distribution shifts. Then, $\mathbf{X}$ is divided into overlapping patches of length $L_p$. The total count of patches, $P$, is calculated as $P = [(T - L_p)/S] + 2$, with $S$ as the sliding stride. These patches, $\mathbf{X}_P \in \mathbb{R}^{L_p \times P}$, are subsequently transformed into a higher-dimensional space, $\hat{\mathbf{X}}_P \in \mathbb{R}^{P \times d_m}$, using a linear transformation where $d_m$ represents the embedding dimension.
\begin{figure}[tbh]
    \centering
    \includegraphics[width=.475\textwidth]{ 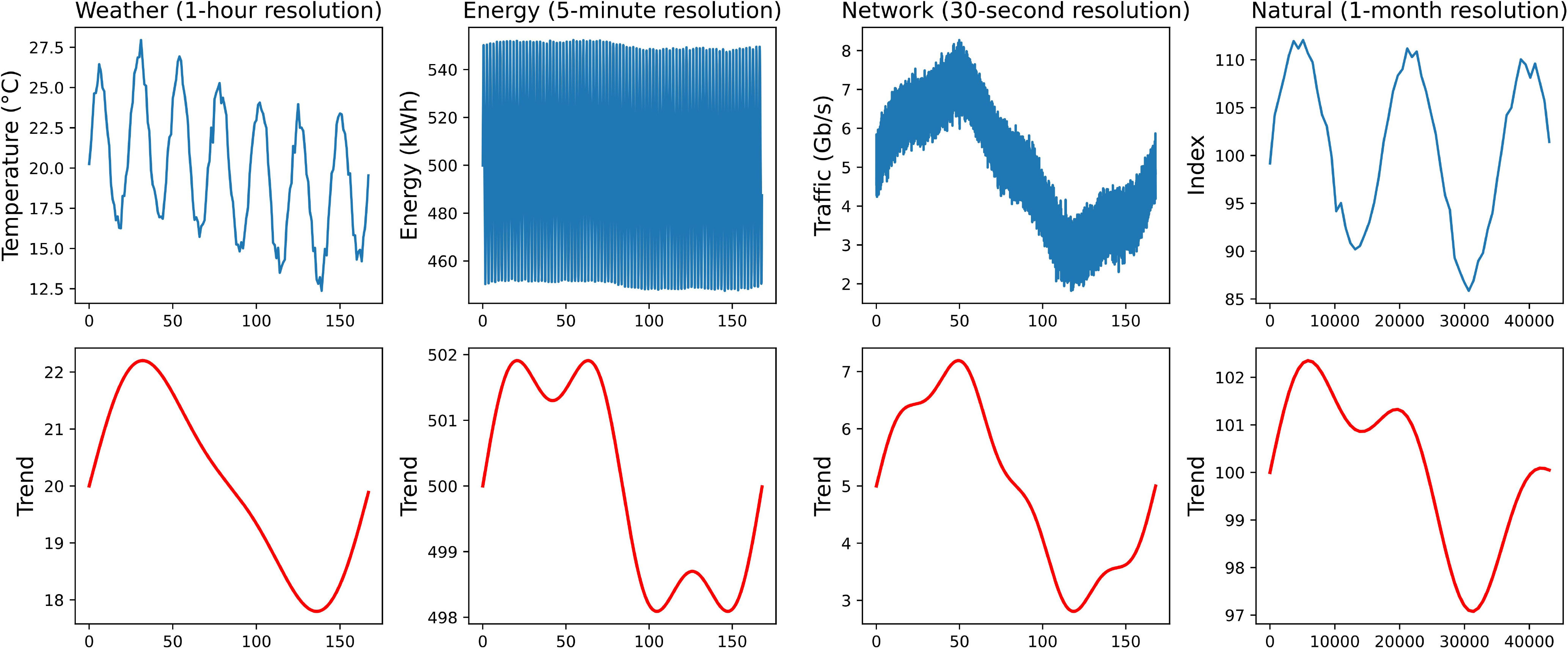}
    \caption{\small Visualization of cross-domain trend similarity within historical observations. \textbf{Upper:} Weather (1-hour resolution), Energy (5-minute resolution), Network (30-second resolution), Natural (1-day resolution), \textbf{Bottom:} Corresponding trend.}
    \label{fig:timescale_pattern}
\end{figure}
\paragraph{Adaptive Trend-awareness Module} Heterogeneous time series can exhibit similar temporal patterns across different timescales, as shown in \textbf{Fig.~\ref{fig:timescale_pattern}}, yet these cross-domain similarities do not typically benefit model training (as discussed in \textbf{Fig.~\ref{fig:example_hete}}). This is means that such time series often contain complex, intertwined temporal patterns. To address this, we propose decoupling individual time series into multiple timescale representations and then applying targeted processing to enhance the model's comprehension of these patterns. This approach aims to improve local model updates and facilitate the sharing of common knowledge during global optimization. Specifically, we introduce Adaptive Trend-awareness Module (ATM) (\textbf{Fig.~\ref{fig:local_architecture}b}), which approaches pattern extraction as a multi-task problem, employing independent operations to mine representations across various timescales. Structurally, ATM comprise a gating network (\textbf{Fig.~\ref{fig:local_architecture}c}), four independent timescale-aware layers—each serving as an expert in extracting representations according to common timescale criteria (e.g., second, minute, hour, day)—and a fusion layer for integrating these representations. The gating network initially decomposes the training sample's representation $\bar{\mX}$ into trend component $\bar{\mX}_{\text{trend}}$ and seasonal component $\bar{\mX}_{\text{seasonal}}$ to form intermediate representations $\mX_{mid}$. Subsequently, it calculates the timescale weight $W$ for expert computation. This process can be described mathematically as follows:
\begin{equation}
\begin{aligned}
    \mX_{mid} = \mW_t \cdot \bar{\mX}_{\text{trend}} &+ \mW_s \cdot \bar{\mX}_{\text{seasonal}}, \\
    W =\texttt{Softmax} &(\text{FFN}(\mX_{mid})).
\end{aligned}
\end{equation}
The top-$k$ weighted experts process the training sample to minimize computational resource consumption. Their predictions are then combined, using these weights, to generate the final output, which is expressed mathematically as:
\begin{equation}
\begin{aligned}
    &\hat{\mX} = \text{FFN}\left[\sum_i W_i \cdot f_i(\mX_{rep})\right] \\
    &\text{where }i \in \{second, hour, minute, day\}
\end{aligned}
\end{equation}
where $\mathbf{X}_{\text{rep}}$ and $\hat{\mathbf{X}}$ signify the latent and predicted representations, respectively. $f(\cdot)$ represents the linear transformation for each timescale, and FFN is the fusion layer. The underlying motivations are: (1) improving the local model updating via treating time series sample modeling into learning multiple potential representation across time scales and (2) facilitating global common knowledge sharing by mining multiple temporal patterns in heterogeneous time series.

\section{FFTS: Optimization}
This section introduces the optimization process of FFTS. Specifically, we implement a unified masking strategy in the local training process to learn sequential patterns of time series and a heterogeneous knowledge alignment strategy in the update process of local and global models to mitigate the impact of heterogeneity on the global model. Furthermore, we apply the global model obtained by FFTS to various time series analysis tasks through a unified downstream adaptation structure. We will elaborate on the details of each component in the following sections.

\paragraph{Unified Masking Strategy} Time series from various domains display significant heterogeneity: simpler patterns can accelerate convergence and simplify representation, whereas complex patterns may slow these processes. This diversity leads clients to learn distinctly different patterns, causing the global model to deviate unpredictably and impeding the development of robust, generalized criteria. To mitigate this, we adopt a uniform masked strategy for each client aimed at learning time series trend representations rather than memorizing domain-specific patterns. Specifically, we mask selected input points in each client's time series and require the model to predict these during training. Binary noise masks $\mM \in \{0, 1\}$ are generated for each sample and applied by element-wise multiplication with the data. We define key parameters, $[L_{\text{m}}, r_m]$, to control the total length of masking and the masking probability. Following~\cite{zerveas2021transformer}, the state transition probabilities are used so that the length of each masked segment adheres to a geometric distribution with a mean of $L_{\text{seg}} = [(1-r_m)/r_m \cdot L_{\text{m}}]$. This strategy focuses the model on analyzing the relationships between consecutive segments, enhancing its capability to identify underlying trends through contextual analysis rather than merely memorizing isolated points.

\paragraph{Heterogeneous Knowledge Alignment} The primary objective of FFTS is to extract common knowledge from heterogeneous time series, thereby establishing a global TSFM with uniform standards. To mitigate the bias introduced by heterogeneity across domains and improve the model performance, we adopt a heterogeneous knowledge alignment strategy into optimization objectives. This involves incorporating ATM-specific regularization terms during both local updates and global optimization to minimize discrepancies between local and global knowledge at various timescales. The local updating process is an unsupervised masked time points reconstruction task. The local optimization objective include MSE for reconstruction accuracy and an additional regularization term specifically for ATM that used for heterogeneous knowledge alignment, can be formulated as:
\begin{equation}
    \gL = \frac{1}{|\mM|}\sum_{t\in \mM} \left[\hat{\mX}(t) - \mX(t)\right]^2 + \lambda || \Theta_{T} - \hat{\Theta}_{T}||^2,
\end{equation}
where $|\mM|$ is the total number of masked time points, $\hat{\mX}(t)$ and $\mX(t)$ represent the prediction of masking points and ground truth, respectively, $\Theta_T$ and $\hat{\Theta}_T$ denote the local ATM's parameters and global ATM's parameters. The global optimization objective can be formulated as:
\begin{equation}
    F(\theta) \text{:}= \argmin \sum_{i=1}^{N} \frac{n_k}{n} \gL_i(\theta_i; D_i) + \lambda || \Theta_{T} - \hat{\Theta}_{T}||^2,
\end{equation}
where $n_i$ and $n$ is the number of samples held by the $k$-th client and all clients, respectively. The motivation is to enhance insights into local and global shared knowledge, integrating local features while maintaining global consistency to improve the adaptability of the global model.

\begin{figure}[tbh]
    \centering
    \includegraphics[width=.48\textwidth]{ 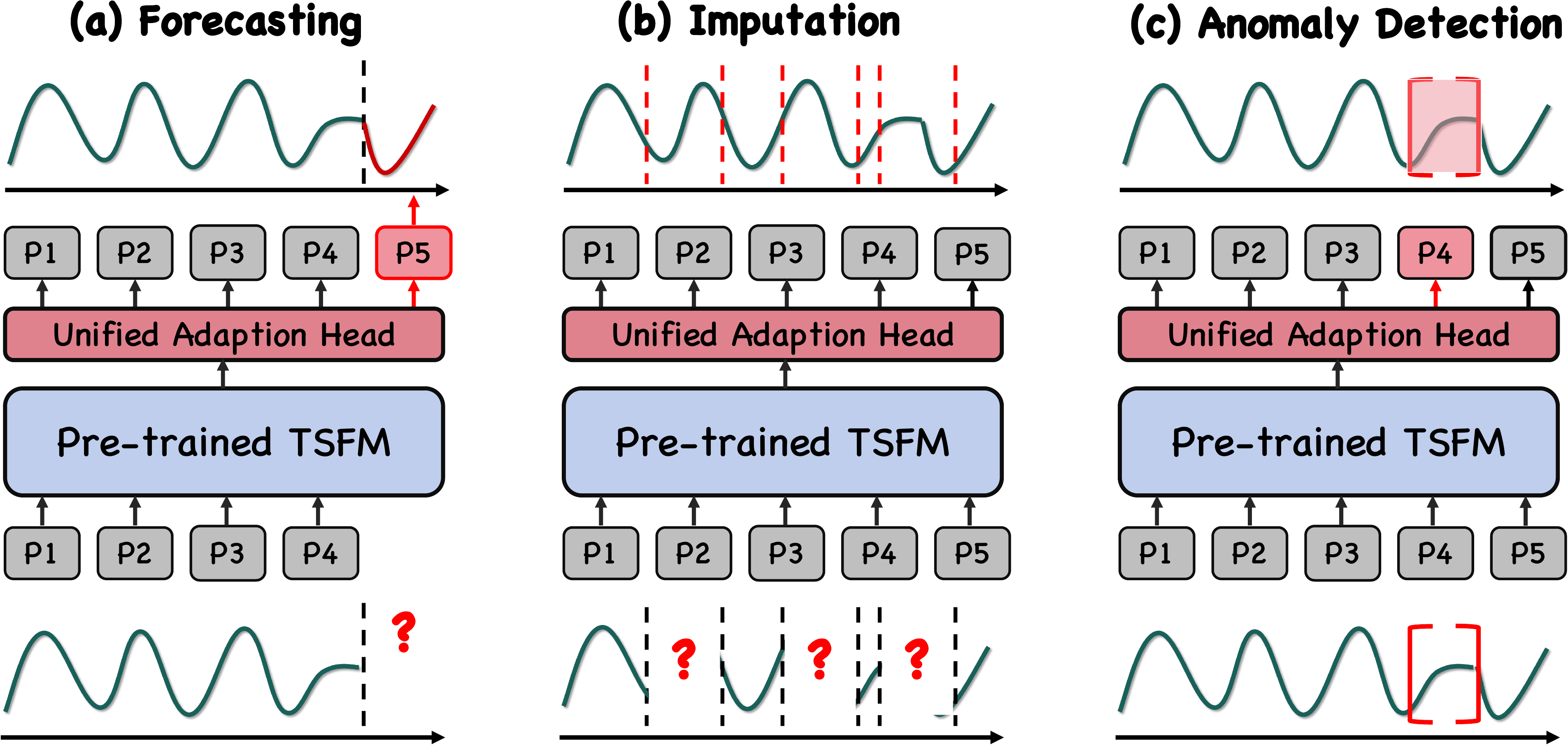}
    \caption{\small Schematic diagram of FFTS for downstream adaption. A unified adaptation head facilitates knowledge transfer across different downstream tasks. (a) Predicting future trends from past data. (b) Filling gaps in data using related time series and context. (c) Identifying unusual patterns in time series.}
    \label{fig:task_illustration}
\end{figure}

\paragraph{Unified Downstream Adaption Architecture} FFTS adapts to downstream time series analysis tasks using a unified architecture (see \textbf{Fig.~\ref{fig:task_illustration}}). This architecture consists of a unified adaptation head, which includes a multi-layer perceptron and layer normalization, enabling efficient processing of downstream tasks with minimal effort.

\section{Experiments}
This section presents our main experimental results, highlighting our FFTS's performance during federated pre-training and its competitive results compared to DL-based time series analysis in different downstream tasks. These tasks include forecasting (long/short-term), interpolation, and anomaly detection. All downstream datasets were excluded from the pre-training phase to prevent data leakage. 

\subsection{Federated Pretraining}
\paragraph{Setups} Federated pre-training is the most critical step in acquiring TSFM in our work. To evaluate the effectiveness of FFTS, we evaluated its performance on a range of different training settings. Specifically, we empirically selected 18 time series datasets (see \textbf{Appendix} for details) from different domains/sources to complete this process and set each dataset to correspond to an independent client. We adopted a uniform pre-training setting of $L = 512$ for each client, and $L_{m} \in \{8, 16, 24\}$ and $r_{m} \in \{15\%, 25\%, 50\%\}$ and $k=3$ for evaluate its performance across different setting. Moreover, to further evaluate the performance of FFTS in FL perspective, we compare it with FL baseline such as FedAvg~\cite{mcmahan2017communication}, FedProx~\cite{li2020federated}, and pFedMe~\cite{t2020personalized}. We also introduced a centralized training strategy, FFTS-Cen, using the same model to demonstrate the effectiveness of FFTS in pre-training for TSFM. During the evaluation phase, we used a uniform configuration with $[L_m=16,r_m=35\%]$. More about pretraining can be found at \textbf{Appendix B}.

\paragraph{Main Results} The main results of pre-training are shown in \textbf{Table~\ref{tab:main_results_pretraining}}. For the various sub-tables: \textbf{(a)} FFTS outperforms baselines and indicate that heterogeneous FL algorithms are not effective as on natural image tasks when dealing with heterogeneous time series. \textbf{(b)} the performance of FFTS is affected by $k$ in the ATM and optimal at $k=3$, which implies that these are indeed multiple cross-timescale trend similarities between heterogeneous time series and that FFTS is able to handle effectively. \textbf{(c)} the effect of client participation rate (PRTP), empirical results show that the PRTP is proportional to the performance. \textbf{(d)} the effect of the regularization weights $\lambda$ associated with ATM on performance. \textbf{(e)} for different timescales layer, the ablation of the trend awareness layer proves the effectiveness of the ATM.
\begin{table}[ht]
    \centering
    \begin{subtable}[t]{0.495\textwidth}
        \centering
        \resizebox{0.95\textwidth}{!}{
        \begin{tabular}{cccc|ccc|rrr|r}
        \toprule
        Length & \multicolumn{3}{c|}{$L_m = 8$} & \multicolumn{3}{c|}{$L_m = 16$} & \multicolumn{3}{c}{$L_m = 24$} &  \multicolumn{1}{|c}{Avg.} \\
        \midrule
        Method / Ratio & 15\%  & 25\%  & 50\%  & 15\%  & 25\%  & 50\%  & \multicolumn{1}{c}{15\%} & \multicolumn{1}{c}{25\%} & \multicolumn{1}{c}{50\%} & \multicolumn{1}{|c}{-} \\
        \midrule
        FedAvg &   \underline{.457}    &   \underline{.449}    &  \underline{.443}   &   \underline{.431}    &   \underline{.448}    &   \underline{.447}    &  \underline{.455}     &   \underline{.435}    & \underline{.437} & .448 \\
        
        FedProx & .633 &   .644    &   .627   &    .673   &   .686    &   .680    & .680      &   .665    & .669 & .662\\
        pFedMe & .613 &   .612    &   .606    &    .654   &   .672    & .665  &    .667   &   .651    &  .642 & .642 \\
        \midrule
        FFTS-Cen &  .433 & .419 & .420 & .417 & .430 & .429 & .442 & .405 & .402 &  \underline{.422}\\
        \bf FFTS (Ours) & \bf .431 & \bf .423 &   \bf .421 & \bf .413 & \bf .421 &   \bf .425    & \bf .436 & \bf .398 & \bf .399 & \bf .418\\
        \bottomrule
        \end{tabular}}
        \caption{Results of FFTS and FL baseline in the 
                pretraining process.}
    \end{subtable}
    \hfill
    \vspace{2pt}
    \begin{subtable}[t]{0.495\textwidth} \small
        \centering
        \begin{subtable}[t]{0.45\textwidth}
            \centering
            \resizebox{\textwidth}{!}{
            \begin{tabular}{cccc}
                \toprule
                Value & $L_m=8$ & $L_m=16$ & $L_m=24$ \\
                \midrule
                $k=1$ & .427 & .432 & \underline{.404}\\
                $k=2$ & .424 & \underline{.427} & .414 \\
                $k=3$ & \bf .423  & \bf .421 & \bf .398\\
                $k=4$ & .456 & .445 & .420\\
                \bottomrule
            \end{tabular}}
            \caption{Hyper-parameter sensitivity of $k$ under 25\% masking ratio.}
        \end{subtable}
        \hfill
        \begin{subtable}[t]{0.495\textwidth}
            \centering
            \resizebox{\textwidth}{!}{
            \begin{tabular}{cccc}
                \toprule
                PRTP & $L_m=8$ & $L_m=16$ & $L_m=24$ \\
                \midrule
                $r_p=10\%$ & .486 & .500 & .532\\
                $r_p=20\%$ & .423  & .439 & .451\\
                $r_p=30\%$ & \underline{.433} & \underline{.422} & \underline{.429}\\
                $r_p=50\%$ & \bf .423 & \bf .421 & \bf .398\\
                \bottomrule
            \end{tabular}}
            \caption{Impact of participation rate on FedTSF with $25\%$ masking ratio.}
        \end{subtable}
        \hfill
        \vspace{2pt}
        \begin{subtable}[t]{0.485\textwidth}
            \centering
            \resizebox{\textwidth}{!}{
            \begin{tabular}{cccc}
                \toprule
                Reg. & $L_m=8$ & $L_m=16$ & $L_m=24$ \\
                \midrule
                $\lambda=5e^{-2}$ &  \underline{.429} & \underline{.430} & \underline{.412}\\
                $\lambda=1e^{-1}$ & \bf .423  & \bf .421 & \bf .398\\
                $\lambda=2e^{-1}$ & .442 & .437 & .420\\
                \bottomrule
            \end{tabular}}
            \caption{Impact of $\lambda$ on $r_m=25\%$ and $k=3$ across various $L_m$ values, $\lambda=1e^{-1}$ is the default.}
        \end{subtable}
        \hfill
        \vspace{2pt}
        \begin{subtable}[t]{0.49\textwidth}
            \centering
            \resizebox{.9\textwidth}{!}{
            \begin{tabular}{cccc}
                \toprule
                Reg. & $L_m=8$ & $L_m=16$ & $L_m=24$ \\
                \midrule
                \textit{wo} day & \bf .426  & \underline{.427} & \underline{.411}\\
                \textit{wo} minute  &  \underline{.427}  & .431 & .416 \\
                \textit{wo} hour & .429  & .430 & .414 \\
                \textit{wo} second & \bf .426 & \bf .425 & \bf .407\\
                \bottomrule
            \end{tabular}}
            \caption{Ablation on ATM across various $L_m$, \textit{wo} means without the corresponding layer.}
        \end{subtable}
    \end{subtable}
    \caption{Main results of federated pre-training process (MSE report). Leading zeros are omitted for values less than one. \textbf{Bold}: the best, \underline{Underline}: the second best.}
    \label{tab:main_results_pretraining}
\end{table}

\paragraph{Framework Ablation Results} We conducted addtional ablation experiments to analyze the impact of specific components within FFTS. These included FFTS without ATM (denoted as FFTS-A,  equivalent to vanilla FedAvg since ATM is a prior of Heterogeneous Knowledge Alignment) and FFTS without Heterogeneous Knowledege Alignment (denoted as FFTS-B). The results shown in \textbf{Table~\ref{tab:framework_ablation}} indicate the effectiveness of the proposed ATM and heterogeneous knowledge alignment strategy.
\begin{table}[tbh]
  \centering
  \resizebox{.4\textwidth}{!}{
    \begin{tabular}{cccccc}
    \toprule
    Method & $L_m=8$ & $L_m=16$& $L_m=24$ & Avg. & Average Var.\\
    \midrule
    FFTS-A &.449 & .448 & .435 & .444 & \bf $\downarrow 7.25\%$ \\
    FFTS-B &.437 & .435 & .420 &  .431 &  \bf $\downarrow 4.11\%$ \\
    \midrule
    FFTS & .423 & .421 & .398 & .414 & - \\
    \bottomrule
    \end{tabular}}
    \caption{\small Ablation results from framework perspective (MSE report). ``Average Var.`` denotes the average variation in performance.}
  \label{tab:framework_ablation}%
  \vspace{-10pt}
\end{table}%

\paragraph{Analysis for Effectiveness of ATM} To further demonstrate the effectiveness of the proposed ATM, we visualized the timescale weights (\textbf{Fig. 6a}) and outputs from the timescale-aware layers (\textbf{Fig. 6b}). The varied timescale weights across different clients indicate that the ATM can adapt its strategy to effectively use different layers depending on the time series data. This method efficiently identifies trend similarities in heterogeneous time series, improving the global model's insights. Moreover, regular outputs from the ATM's four timescale-aware layers under sinusoidal functions confirm their capability to detect trends across timescales, validating the ATM's effectiveness.

\begin{figure}[tbh]
    \centering
    \includegraphics[width=.485\textwidth]{ 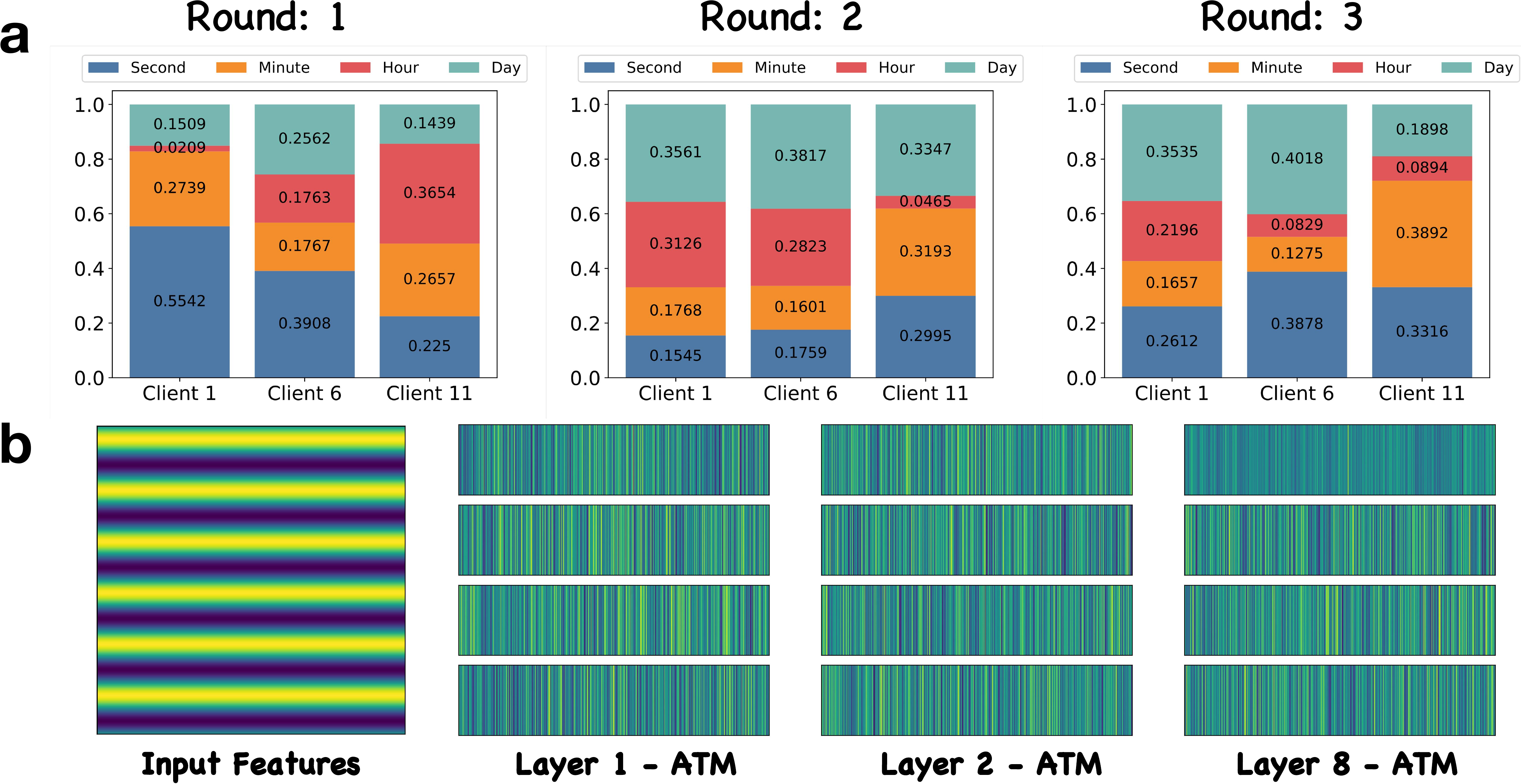}
    \caption{\small Visualization of our proposed ATM. \textbf{(a)}: variations in timescale weights for selected clients across various communication rounds. \textbf{(b)}: featuring regular sinusoidal patterns for the rightmost input features. Subsequent depict the Gating network representation in the ATM's first, second, and final layers. From top to bottom, represent second-, minute-, hour-, and day-aware layers.}
    \label{fig:atm_vis}
\end{figure}

\subsection{Downstream Baseline}
We evaluated the performance of our FFTS on various downstream tasks, including forecasting, imputation, and anomaly detection, using a widely adopted DL-based time series analysis model following~\cite{zhou2023one, jin2023time}. We also compared FFTS with a well-known FL baseline to assess the effectiveness of TSFMs trained with FL algorithms. \textbf{Table \ref{tab:baseline_information}} provides an overview of the main baselines and their application tasks, with detailed descriptions available in \textbf{Appendix B}.

\begin{table}[tbh]  
  \centering
  \resizebox{.48\textwidth}{!}{
    \begin{tabular}{c|c|c}
    \toprule
    Baseline & Venue & Downstream Tasks \\
    \midrule
    LogTransformer~\cite{li2019enhancing} &    NeurIPS 2019   &  Anomaly Detection\\
    N-BEATS~\cite{oreshkin2020nbeatsneuralbasisexpansion} & arXiv 2019 &  Forecasting \\
    Reformer~\cite{kitaev2020reformer} &  ICLR 2020   & Forecasting, Imputation, Anomaly Detection \\
    Informer~\cite{zhou2021informer} &   AAAI 2021    &  Forecasting, Imputation, Anomaly Detection\\
    LightTS~\cite{zhang2022less} &  arXiv 2022   &  Forecasting, Imputation\\
    ETSformer~\cite{woo2022etsformer} &  arXiv 2022     &  Forecasting, Imputation, Anomaly Detection\\
    Stationary~\cite{liu2022non} &  NeurIPS 2023     & Forecasting, Imputation, Anomaly Detection \\
    Autoformer~\cite{wu2021autoformer} &   NeurIPS 2021    & Forecasting, Imputation, Anomaly Detection \\
    FEDformer~\cite{zhou2022fedformer} &  ICML 2022     & Forecasting, Imputation, Anomaly Detection \\
    Pyraformer~\cite{liu2021pyraformer} &  ICLR 2023   & Anomaly Detection \\
    AnomalyTransformer~\cite{xu2022anomalytransformertimeseries} &  ICLR 2022     & Anomaly Detection \\
    TimesNet~\cite{wu2022timesnet} &   ICLR 2023    & Forecasting, Imputation, Anomaly Detection \\
    PatchTST~\cite{nie2022time}&  ICLR 2023     & Forecasting, Imputation, Anomaly Detection \\
    DLinear~\cite{zeng2023transformers} &   AAAI 2023    & Forecasting, Imputation, Anomaly Detection \\
    N-HiTS~\cite{challu2022nhitsneuralhierarchicalinterpolation} &  AAAI 2023 & Forecasting \\
    GPT4TS~\cite{zhou2023one} &   NeurIPS 2023    &  Forecasting, Imputation, Anomaly Detection\\
    LLMTime~\cite{gruver2024large}  & NeurIPS 2023 & Long-term Forecasting (zero-shot only) \\
    Time-LLM~\cite{jin2023time} &   ICLR 2024    & Forecasting \\
    \midrule
    FedAvg*~\cite{mcmahan2017communication} & AISTATS 2017  &  Forecasting, Imputation, Anomaly Detection \\
    FedProx*~\cite{li2020federated} & MLSys 2020 & Forecasting, Imputation, Anomaly Detection \\
    pFedMe*~\cite{t2020personalized} & NeurIPS 2020 & Forecasting, Imputation, Anomaly Detection \\
    FFTS* & This paper & Forecasting, Imputation, Anomaly Detection \\
    \bottomrule
    \end{tabular}}
    \caption{\small Description of deep time series analysis baseline used in the downstream time series analysis task. `*` means that a TSFM trained by the corresponding FL algorithm.}
    \label{tab:baseline_information}%
\end{table}%

\subsection{Time Series Forecasting}
\paragraph{Setups} Time series forecasting is essential yet challenging in real-world applications. To evaluate performance, we adopt popular benchmarks and experimental setting following~\cite{jin2023time}, including ETT (ETTh1, ETTh2, ETTm1, ETTm2), Weather, and Illness datasets (excluding Traffic and Electricity due to their presence in pretraining), with a unified look-back window length of 512. We conduct three types of downstream experiments: (1) Regular: fine-tuning the pretrained TSFM before evaluation; (2) Few-shot: fine-tuning the pretrained TSFM using 5\% or 10\% of data; and (3) Zero-shot: evaluating the pretrained TSFM without training on target datasets (only ETT-series dataset).

\begin{table}[tbh]
  \centering
  \resizebox{.47\textwidth}{!}{
    \begin{tabular}{cccccccc}
    \toprule
    Method & ETTh1 & ETTh2 & ETTm1 & ETTm2 & Weather & ILI & Avg.\\
    \midrule
    Reformer & 1.029 & 6.736 & 0.799 & 1.479 & 0.803 & 4.724 & 2.595 \\
    Informer & 1.040  & 4.431 & 0.961 & 1.410  & 0.634 & 5.137 & 2.269 \\
    LightTS & 0.491 & 0.602 & 0.435 & 0.409 & 0.261 & 7.382 & 1.597 \\
    ETSformer & 0.542 & 0.439 & 0.429 & 0.293 & 0.271 & 2.497 & 0.745 \\
    Stationary & 0.570  & 0.526 & 0.481 & 0.306 & 0.288 & 2.077 & 0.708 \\
    Autoformer & 0.496 & 0.450  & 0.588 & 0.327 & 0.338 & 3.006 & 0.868 \\
    FEDformer & 0.440  & 0.437 & 0.448 & 0.305 & 0.309 & 2.847 & 0.798 \\
    TimesNet & 0.458 & 0.414 & 0.400   & 0.291 & 0.259 & 2.139 & 0.660 \\
    PatchTST & 0.413 & \underline{0.330}  & 0.351 & 0.255 & 0.225 & 1.443 & 0.503 \\
    Dlinear & 0.422 & 0.431 & 0.357 & 0.267 & 0.248 & 2.169 & 0.649 \\
    GPT4TS & 0.465 & 0.381 & 0.388 & 0.284 & 0.237 & 1.925 & 0.613\\
    Time-LLM & 0.408 & 0.334 & 0.329 & \underline{0.251} & 0.225 & 1.435 & 0.497\\
    \midrule
    FedAvg (FT) & 0.400 & 0.342 & 0.335&  0.251& 0.218& 1.390& 0.489\\
    FedProx (FT) & 0.425 & 0.354& 0.370& 0.309 & 0.263& 1.550& 0.545\\
    pFedMe (FT)& 0.425& 0.353& 0.372& 0.302 & 0.260& 1.553& 0.544\\
    \midrule
    FFTS-Cen (FT) & 0.392 & 0.331 &  \bf 0.314 &  0.245 & \underline{0.217}  & \underline{1.353} & \underline{0.475}\\
    \bf FFTS (Ours, FT) & \bf 0.389 & \bf 0.329 & \underline{0.315} & \bf 0.244 & \bf 0.215 & \bf 1.348 & \bf 0.473\\
    \bottomrule
    \end{tabular}}
    \caption{\small Long-term forecasting results average across prediction horizons $\{96, 192, 336, 720\}$ and $\{24, 36, 48, 60\}$ for ILI dataset. \textbf{Bold}: the best, \underline{Underline}: the second best. (FT) denotes the obtained TSFM with fine-tuning on corresponding datasets. \textbf{Appendix D} shows the full results.}
  \label{tab:long_term_forecasting_simple}%
  \vspace{-4pt}
\end{table}%

\paragraph{Main Results} \textbf{Table~\ref{tab:long_term_forecasting_simple}} demonstrates the TSFM obtained by our FFTS can achieve best long-term forecasting performance, surpassing the SOTA method Time-LLM by averaged \textbf{9.52\%} across six datasets. Notably, the TSFM obtained by vanilla FedAvd can also achieve very competitive performance compared to Time-LLM, with a difference of only \textbf{1.7\%}. These indicate that (1) FL can be used as an effective pre-training strategy for unimodal TSFM, and (2) our FFTS is more effective in dealing with heterogeneous temporal pre-training compared to comparable baselines.

\paragraph{Few-/Zero-shot Results} Few/zero-shot generalization are crucial capabilities for TSFMs. \textbf{Table~\ref{tab:few_shot_forecasting_simple}} illustrates the few-shot generalization capability of FFTS, which outperforms baseline models. FFTS achieves optimal performance, surpassing Time-LLM by an average of \textbf{4.1\%} with limited training data. Notably, the TSFM obtained through FedAvg also exhibits superior performance, showing an average \textbf{1.1\%} improvement over Time-LLM. Despite its comparatively lower pretraining performance, achieves competitive results against Time-LLM. \textbf{Table~\ref{tab:zero_shot_forecsating_simple}} presents zero-shot results on the ETT dataset, where the TSFM obtained by our FFTS outperforms Time-LLM by an average of \textbf{3.87\%}. Most significant, FFTS outperforms the centralised training baseline (FFTS-Cen) on the same architecture and pretraining datasets in most case. These results underscore the comparable performance of our model to pre-trained LLMs, even with limited training datasets, and underscore the effectiveness and superiority of the proposed FFTS.
\begin{table}[tbh]\small
  \centering
  \resizebox{.485\textwidth}{!}{
    \begin{tabular}{ccc|cc|cc|cc|cc|c}
    \toprule
    Method & \multicolumn{2}{c|}{ETTh1} & \multicolumn{2}{c|}{ETTh2} & \multicolumn{2}{c|}{ETTm1} & \multicolumn{2}{c|}{ETTm2} & \multicolumn{2}{c|}{Weather} & \multirow{2}[4]{*}{Avg.} \\
\cmidrule{1-11}    Ratio & 10\%  & 5\%   & 10\%  & 5\%   & 10\%  & 5\%   & 10\%  & 5\%   & 10\%  & 5\%   &  \\
    \midrule
    Reformer & 1.249 & 1.241 & 3.485 & 3.527 & 1.426 & 1.264 & 3.978 & 3.581 & 0.546 & 0.447 & 2.074 \\
    Informer & 1.199 & 1.225 & 3.872 & 3.922 & 1.192 & 1.163 & 3.37  & 3.658 & 0.597 & 0.584 & 2.078 \\
    LightTS & 1.375 & 1.451 & 2.655 & 3.206 & 0.971 & 1.123 & 0.987 & 1.415 & 0.289 & 0.305 &  1.378\\
    ETSformer & 1.180  & 1.189 & 0.894 & 0.809 & 0.980  & 1.125 & 0.447 & 0.534 & 0.318 & 0.333 &  0.781\\
    Stationary & 0.915 & 0.943 & 0.462 & 0.470  & 0.797 & 0.857 & 0.332 & 0.341 & 0.318 & 0.327 &  0.576\\
    Autoformer & 0.702 & 0.722 & 0.488 & 0.441 & 0.802 & 0.796 & 1.342 & 0.388 & 0.300   & 0.310  &  0.629\\
    FEDformer & 0.639 & 0.658 & 0.466 & 0.463 & 0.722 & 0.730  & 0.463 & 0.381 & 0.284 & 0.309 &  0.512\\
    TimesNet & 0.869 & 0.925 & 0.479 & 0.439 & 0.677 & 0.717 & 0.320  & 0.344 & 0.279 & 0.298 &  0.535\\
    PatchTST & 0.633 & 0.694 & 0.415 & 0.827 & 0.501 & 0.526 & 0.296 & 0.314 & 0.242 & 0.269 &  0.472\\
    DLinear & 0.691 & 0.750  & 0.605 & 0.694 & 0.411 & \bf 0.400   & 0.316 & 0.399 & 0.241 & 0.263 & 0.477 \\
    GPT4TS & 0.590  & 0.681 & 0.397 & 0.400   & 0.464 & 0.472 & 0.293 & 0.308 & 0.238 & 0.263 & 0.411 \\
    Time-LLM & \underline{0.556} & \underline{0.627} & \underline{0.370}  & \underline{0.382} & \underline{0.404} & 0.425 & \underline{0.277} & \underline{0.274} & \underline{0.234} & \underline{0.260}  &  0.381\\
    \midrule
     FedAvg (FT)& 0.562 &  0.618 & 0.369
     & 0.371  &  0.398& 0.420 & 0.275 &  0.278 & 0.227&  0.256&  0.377\\
    FedProx (FT)& 0.577 & 0.630 & 0.382
     & 0.384 & 0.414 &  0.431 & 0.285 & 0.299& 0.233 & 0.269 &  0.390\\
    pFedMe (FT)& 0.575 & 0.629 & 0.374
    &  0.380& 0.421 &  0.433 & 0.290& 0.295 & 0.230 & 0.262  &  0.389\\
    \midrule
    FFTS-Cen (FT) & 0.549 & 0.609 & 0.363 & 0.365 & 0.387 & 0.417 & 0.270 & 0.263 & 0.219 & 0.247 & \underline{0.369}\\
    FFTS (Ours, FT) & \bf 0.546 & \bf 0.606 & 
    \bf 0.362 & \bf 0.363 & \bf 0.386 & \underline{0.411} & \bf 0.266 & \bf 0.259 & \bf 0.216 & \bf 0.245 &  \bf 0.366\\
    \bottomrule
    \end{tabular}}
      \caption{\small Few-shot results on long-term forecasting average across prediction horizons of $\{96, 192, 336, 720\}$ with ratio $\{10\%, 5\%\}$ of training dataset. \textbf{Bold}: the best, \underline{Underline}: the second best. (FT) denotes the obtained TSFM with fine-tuning on corresponding datasets. Full results are in \textbf{Appendix D}.}
  \label{tab:few_shot_forecasting_simple}%
\end{table}%

\begin{table}[tbh]
\vspace{-10pt}
  \centering
  \resizebox{.478\textwidth}{!}{
    \begin{tabular}{ccc|ccccccc}
    \toprule
    Methods & FFTS (FT) & FFTS-Cen (FT) & Time-LLM & LLMTime & GPT4TS & DLinear & PatchTST & TimesNet & Autoformer \\
    \midrule
    h1$\Rightarrow$h2 & \bf 0.339 & \underline{0.343} &\underline{0.353} & 0.992 & 0.406 & 0.493 & 0.380  & 0.421 & 0.582 \\
    h1$\Rightarrow$m2 & \bf 0.266 & \underline{0.269} &\underline{0.273} & 1.867 & 0.325 & 0.415 & 0.314 & 0.327 & 0.457 \\
    \midrule
    h2$\Rightarrow$h1 & \bf 0.460  & \underline{0.463} &\underline{0.479} & 1.961 & 0.757 & 0.703 & 0.565 & 0.865 & 0.757 \\
    h2$\Rightarrow$m2 & \bf 0.263 & \bf 0.263 &\underline{0.272} & 1.867 & 0.335 & 0.328 & 0.325 & 0.342 & 0.366 \\
    \midrule
    m1$\Rightarrow$h2 & \underline{0.369} & \bf 0.367 &0.381 & 0.992 & 0.433 & 0.464 & 0.439 & 0.457 & 0.47 \\
    m1$\Rightarrow$m2 & \bf 0.249 & \underline{0.254} &\underline{0.268} & 1.867 & 0.313 & 0.335 & 0.296 & 0.322 & 0.469 \\
    \midrule
    m2$\Rightarrow$h2 & \bf 0.339 & \underline{0.345} &0.354 & 0.992 & 0.435 & 0.455 & 0.409 & 0.435 & 0.423 \\
    m2$\Rightarrow$h1 & \bf 0.403 & \underline{0.405} &0.414 & 1.933 & 0.769 & 0.649 & 0.568 & 0.769 & 0.753 \\
    \midrule
    Avg. & \bf 0.336 & \underline{0.339} &0.349 & 1.559& 0.472 &  0.480 &  0.412 & 0.492 &  0.525\\
    \bottomrule
    \end{tabular}}
  \caption{\small Zero-shot results (MSE report). The source dataset is ETT. \textbf{Bold}: the best, \underline{Underline}: the second best. (FT) denotes the obtained TSFM with fine-tuning on corresponding datasets. $\Rightarrow$ means the transfer of the source dataset to the target dataset (ETT series). Full results are in \textbf{Appendix D}.}
\label{tab:zero_shot_forecsating_simple}
\end{table}%

\paragraph{Short-term Forecasting Setups \& Results} Short-term forecasting is crucial in real-world applications. We conducted experiments to further evaluate the effectiveness of our FFTS using the M4 dataset. \textbf{Table~\ref{tab:short_term_forecasting_simple}} shows that our FFTS outperforms advanced time series models, including LLM-based Time-LLM and GPT4TS, further demonstrating FFTS's effectiveness and superiority.
\begin{table}[tbh]
  \centering
  \resizebox{0.25\textwidth}{!}{
    \begin{tabular}{cccc}
    \toprule
    Method & SMAPE & MASE  & OWA \\
    \midrule
    Informer & 14.086 & 2.718 & 1.23 \\
    Autoformer & 12.909 & 1.771 & 0.939 \\
    Stationary & 12.78 & 1.756 & 0.93 \\
    FEDformer & 13.16 & 1.775 & 0.949 \\
    DLinear & 13.639 & 2.095 & 1.051 \\
    LightTS & 13.525 & 2.111 & 1.051 \\
    ETSformer & 14.718 & 2.408 & 1.172 \\
    N-BEATS & 12.25 & 1.698 & 0.896 \\
    N-HiTS & 12.035 & 1.625 & 0.869 \\
    PatchTST & 12.059 & 1.623 & 0.869 \\
    TimesNet & 12.88 & 1.836 & 0.955 \\
    GPT4TS & 12.69 & 1.808 & 0.94 \\
    Time-LLM & 11.983 & 1.595 & 0.859 \\
    \midrule
    FedAvg (FT) & 12.302 & 1.634 & 0.895 \\
    FedProx (FT) & 12.582 & 1.783 & 1.029 \\
    pFedMe (FT)& 12.563 & 1.733 & 0.950 \\
    \midrule
    FFTS-Cen (FT) &  11.937 & 1.569 & 0.880\\
    \bf FFTS (Ours, FT) & \bf 11.664 & \bf 1.556 & \bf 0.868 \\
    \bottomrule
    \end{tabular}}
  \caption{\small Short-term forecasting resuls. The forecasting length are $\{6, 48\}$ an results are weighted averaged from server datasets under different sample intervals. \textbf{Bold}: the best, \underline{Underline}: the second best. Full results are in \textbf{Appendix D}.}
\label{tab:short_term_forecasting_simple}%
\end{table}%

\subsection{Time Series Imputation}
\paragraph{Setups} Imputation aiming to fill corrupted time series based on partially observed data. We conduct experiment on five popular real-world datasets, including ETT (ETTh1, ETTh2, ETTm1, ETTm2), and Weather, where the data-missing is common. Following the experiment setting of GPT4TS~\cite{zhou2023one}, different random mask ratio $\{96, 192, 336, 720\}$ of time points are selected for the evaluation on various proportions of missing data.
\paragraph{Result} \textbf{Table~\ref{tab:imputation_simple}} demonstrates our proposed FFTS can achieve the best performance across different datasets. Compared to the state-of-the-art GPT4TS, the TSFM trained using FFTS demonstrates superior performance, reducing MSE by \textbf{14.7\%}. Furthermore, the TSFM obtained through vanilla FedAvg surpasses GPT4TS by \textbf{2.6\%}. FFTS also achieves competitive performance against the centralised strategy. These results validate both the effectiveness of FL in TSFM pre-training and the superiority of our FFTS.

\begin{table}[tbh]\small
  \centering
  \resizebox{.47\textwidth}{!}{
    \begin{tabular}{ccccccc}
    \toprule
    Method & ETTh1 & ETTh2 & ETTm1 & ETTm2 & Weather & Avg. \\
    \midrule
    Reformer & 0.055 & 0.157 & 0.122 & 0.234 & 0.038 & 0.121 \\
    Informer & 0.071 & 0.156 & 0.161 & 0.337 & 0.045 & 0.154 \\
    LightTS & 0.051 & 0.029 & 0.103 & 0.055 & 0.031 & 0.054 \\
    ETSformer & 0.036 & 0.026 & 0.094 & 0.053 & 0.032 & 0.048 \\
    Stationary & 0.062 & 0.101 & 0.117 & 0.163 & 0.099 & 0.108 \\
    Autoformer & 0.093 & 0.096 & 0.201 & 0.142 & 0.052 & 0.117 \\
    FEDformer & 0.104 & 0.046 & 0.284 & 0.119 & 0.055 & 0.122 \\
    TimesNet & 0.120  & 0.208 & 0.202 & 0.367 & 0.076 & 0.195 \\
    PatchTST & 0.047 & 0.029 & 0.115 & 0.065 & 0.060 & 0.063 \\
    DLinear & 0.027 & 0.022 & 0.078 & 0.049 & 0.030 & 0.041 \\
    GPT4TS & 0.028 & \underline{0.021} & 0.069 & 0.048 & 0.031 & 0.039 \\
    \midrule
    FedAvg (FT) &  0.026 & 0.022 & 0.063 & 0.046 & 0.031 & 0.038 \\
    FedProx (FT) & 0.030  & 0.026 & 0.073 & 0.047& 0.034 & 0.042 \\
    pFedMe (FT) & 0.031 & 0.024 & 0.071 & 0.049 & 0.032& 0.041 \\
    \midrule
    FFTS-Cen (FT) & \underline{0.024} & \bf 0.018 & \bf 0.057 & \underline{0.045  }& \bf 0.029 & \underline{0.035}\\
    FFTS (Ours, FT) & \bf 0.023 & \bf 0.018 & \underline{0.058} & \bf 0.044 & \bf 0.029 & \bf 0.034 \\
    \bottomrule
    \end{tabular}}
    \caption{\small Imputation results. Time points are randomly masked at ratios of $\{12.5\%, 25\%, 37.5\%, 50\%\}$ with an input length of 96. The results presented are averaged across these four different mask ratios.  \textbf{Bold}: the best, \underline{Underline}: the second best. (FT) denotes the obtained TSFM with fine-tuning on corresponding datasets. Full results are available in \textbf{Appendix D}.}
    \label{tab:imputation_simple}
\end{table}%

\subsection{Time Series Anomaly Detection}
\paragraph{Setups} Anomaly detection plays a crucial role in industrial operations. We benchmark FFTS against five widely utilized datasets: SMD, MSL, SMAP, SwaT, and PSM, adhering to the experimental protocols of GPT4TS~\cite{zhou2023one} to guarantee a fair comparison. Detailed information about datasets and experiments are in \textbf{Appendix B/D}.
\begin{table}[tbh]\small
  \centering
  \resizebox{.47\textwidth}{!}{
    \begin{tabular}{ccccccr}
    \toprule
    Method & SMD   & MSL   & SMAP  & SWaT  & PSM   & \multicolumn{1}{c}{Avg F1} \\
    \midrule
    Transformer & 79.56 & 78.68 & 69.70  & 80.37 & 76.07 & \multicolumn{1}{c}{76.88} \\
    LogTransformer & 76.21 & 79.57 & 69.97 & 80.52 & 76.74 & \multicolumn{1}{c}{76.60} \\
    Autoformer & 85.11 & 79.05 & 71.12 & 92.74 & 93.29 & \multicolumn{1}{c}{84.26} \\
    Pyraformer & 83.04 & 84.86 & 71.09 & 91.78 & 82.08 & \multicolumn{1}{c}{82.57} \\
    Informer & 81.65 & 84.06 & 69.92 & 81.43 & 77.10  & \multicolumn{1}{c}{78.83} \\
    Reformer & 75.32 & 84.40  & 70.40  & 82.80  & 73.61 & \multicolumn{1}{c}{77.31} \\
    ETSformer & 83.13 & \bf 85.03 & 69.50  & 84.91 & 91.76 & \multicolumn{1}{c}{82.87} \\
    FEDformer & 85.08 & 78.57 & 70.76 & 93.19 & 97.23 & \multicolumn{1}{c}{84.97} \\
    Stationary & 84.62 & 77.50  & 71.09 & 79.88 & 97.29 & \multicolumn{1}{c}{82.08} \\
    AnomalyTransformer & 85.49 & 83.31 & 71.18 & 83.10  & 79.40  & \multicolumn{1}{c}{80.50} \\
    TimesNet & 84.61 & 81.84 & 69.39 & 93.02 & 97.34 & \multicolumn{1}{c}{85.24} \\
    PatchTST & 84.62 & 78.70  & 68.82 & 85.72 & 96.08 & \multicolumn{1}{c}{82.78} \\
    DLinear & 77.10  & \underline{84.88} & 69.26 & 87.52 & 93.55 & \multicolumn{1}{c}{82.46} \\
    GPT4TS & 86.89 & 82.45 & 72.88 & 94.23 & 97.13 & \multicolumn{1}{c}{86.72} \\
    \midrule
    FedAvg (FT)&   87.21    &   83.33    &   \underline{74.20}   &   \underline{95.00}    &  \underline{98.20}   & \multicolumn{1}{c}{\underline{87.59}} \\
    FedProx (FT)&   85.45    &  82.21     &   70.09    &   92.81    &  96.59     &  \multicolumn{1}{c}{85.43}\\
    pFedMe (FT)&   85.78    &   83.04    &    71.11   &   91.99    &   96.20    &  \multicolumn{1}{c}{85.62}\\
    \midrule
    FFTS-Cen (FT) & 87.93 & 83.56 & 74.08 & 94.94 & 97.95 & \multicolumn{1}{c}{87.69}\\
    \bf FFTS (Ours, FT) &   \bf  88.42   &   84.01    &   \bf 74.53    &  \bf 95.27     &  \bf  98.25    & \multicolumn{1}{c}{\bf 88.10} \\
    \bottomrule
    \end{tabular}}
    \caption{\small Anomaly detection results. We calculate the F1-score (\%) for each dataset and statics the average F1-score. \textbf{Bold}: the best, \underline{Underline}: the second best. \textbf{Appendix D} shows the full results.}
    \label{tab:anomaly_detection_simple}
\end{table}%

\paragraph{Results} \textbf{Table~\ref{tab:anomaly_detection_simple}} reveals that the TSFM developed via our FFTS exhibits superior performance, achieving an average F1-score of \textbf{88.10\%}, which surpasses the previous SOTA GPT4TS by \textbf{1.02\%}. Additionally, the TSFM trained with FFTS outperforms its counterpart trained via FL baseline by an average of \textbf{1\%}. Notably, the TSFM trained via FedAvg exceeds the performance of GPT4TS, underscoring the effectiveness of FL in improving TSFM training and demonstrating the efficacy of FFTS. The competitive performance relative to the centralised training baseline (FFTS-Cen) further indicates the superiority of FFTS.

\section{Discussion}
This section provides additional discussion about the computational and communication efficiency, as well as privacy of our proposed FFTS framework.
\subsection{Computational/Communication Efficiency}
We analyze computational and communication efficiency in terms of complexity, with $N$ representing the number of clients, $d$ the model dimension, $M$ the data volume, and $E$ the number of local training rounds. Computational efficiency: Each ATM expert has a complexity of $\gO(d)$, and a $L$-layer Transformer reaches $\gO(L \cdot d)$. With $k$ experts activated per sample by ATM, the complexity is $\gO(k \cdot d)$. Adding local optimization, the total complexity per client per round is $\gO(E \cdot (L \cdot d^2 + k \cdot d + M \cdot d))$. ATM's MoE structure reduces computational load, though the regularization term adds a minor overhead to enhance model consistency. Communication efficiency: Per-round communication complexity is $\gO(N \cdot D)$, where $D$ represents model parameters. Regularization mitigates frequent global updates, thus lowering bandwidth requirements in federated settings. These mean that our proposed FFTS has a significant potential to support the training of TSFMs in low-resource environments (e.g., meteorological~\cite{chen2023tempee,chen2022dynamic}, healthcare~\cite{chen2023interpretable,ren2024federated}, and science~\cite{chen2023collaborative,ren2024distributed}, etc.).

\subsection{Privacy Guarantee}
Our FFTS enhances privacy through three aspects: (1) Clients retain their data locally, sharing only model parameters during training, which prevents exposure of raw data. (2)  We use additional regularization term to balance model updates, reducing the risk of overfitting to specific client's distributions and protecting sensitive information. (3) ATM selectively activates model components based on the processed data, ensuring only pertinent data influences the federated model and reducing the data footprint in updates.

\section{Conclusion}
This paper demonstrates the potential of FL for training foundation models on heterogeneous time series datasets. We introduce FFTS, a novel FL approach designed to address heterogeneity in time series foundation model training. FFTS considers each data-holding organization as an independent client within a collaborative framework. Each client trains a local model to preserve the unique characteristics of its dataset, while a server aggregates these models to form a time series foundation model. FFTS enhances training through model architecture and optimization, introducing an adaptive trend-awareness module, a uniform masking strategy, and a heterogeneous knowledge alignment strategy. A unified adaptation architecture supports various downstream tasks. Extensive experiments on real-world time series datasets demonstrate FFTS's robust generalization capabilities in forecasting, imputation, and anomaly detection.

\bibliography{aaai25}

\end{document}